%% file: root.tex
\documentclass[letterpaper, 10 pt, conference]{ieeeconf}  

\IEEEoverridecommandlockouts                              

\title{\LARGE \bf
Efficient Reinforcement Learning of Task Planners for Robotic Palletization through Iterative Action Masking Learning}

\author{Zheng Wu$^{1}$,  Yichuan Li$^{2}$, Wei Zhan$^{1}$, Changliu Liu$^{3}$, Yun-Hui Liu$^{2}$, Masayoshi Tomizuka$^{1}$
\thanks{*This work is supported by the InnoHK of the Government of the Hong Kong Special Administrative Region via the Hong Kong Centre for Logistics Robotics. Corresponding to Zheng Wu (e-mail: zheng\_wu@berkeley.edu).}%
\thanks{$^{1}$Zheng Wu, Wei Zhan, Masayoshi Tomizuka are with the Department of Mechanical Engineering, 
		University of California, Berkeley, CA.}
\thanks{$^{2}$Yichuan Li, Yun-Hui Liu are with the T Stone Robotics Institute, the Department of Mechanical and Automation Engineering, the Chinese University of Hong Kong.}
\thanks{$^{3}$Changliu Liu is with the Robotics Institute, Carnegie Mellon University, PA.}
}
\usepackage{graphicx}
\usepackage{xcolor}
\usepackage{mathtools}
\usepackage{algorithm}
\usepackage{algorithmicx}
\usepackage{algpseudocode}
\usepackage{amsmath}
\usepackage[inkscapelatex=false]{svg}
\usepackage{lettrine}
\usepackage{multirow}
\usepackage{booktabs}
\usepackage{cite}
\usepackage{setspace}
\usepackage{bbold}
\usepackage{amsfonts}
\usepackage{hyperref}
\usepackage{graphicx} 
\begin{document}

\maketitle
\thispagestyle{empty}
\pagestyle{empty}

\input{abstract}
\input{indexterm}
\input{introduction}
\input{related}
\input{framework}
\input{experiment}

\input{conclusion}

\bibliographystyle{IEEEtran}
\bibliography{ref}

\end{document}

%% file: abstract.tex
\begin{abstract}
The development of robotic systems for palletization in logistics scenarios is of paramount importance, addressing critical efficiency and precision demands in supply chain management. 
This paper investigates the application of Reinforcement Learning (RL) in enhancing task planning for such robotic systems. 
Confronted with the substantial challenge of a vast action space, which is a significant impediment to efficiently apply out-of-the-shelf RL methods, our study introduces a novel method of utilizing supervised learning to iteratively prune and manage the action space effectively. 
By reducing the complexity of the action space, our approach not only accelerates the learning phase but also ensures the effectiveness and reliability of the task planning in robotic palletization. 
The experiemental results underscore the efficacy of this method, highlighting its potential in improving the performance of RL applications in complex and high-dimensional environments like logistics palletization. 
\end{abstract}

%% file: indexterm.tex
\vspace{0.5em}

%% file: introduction.tex
\section{INTRODUCTION}
\begin{figure}[t]
	\centering
	\includegraphics[width=0.95\linewidth]{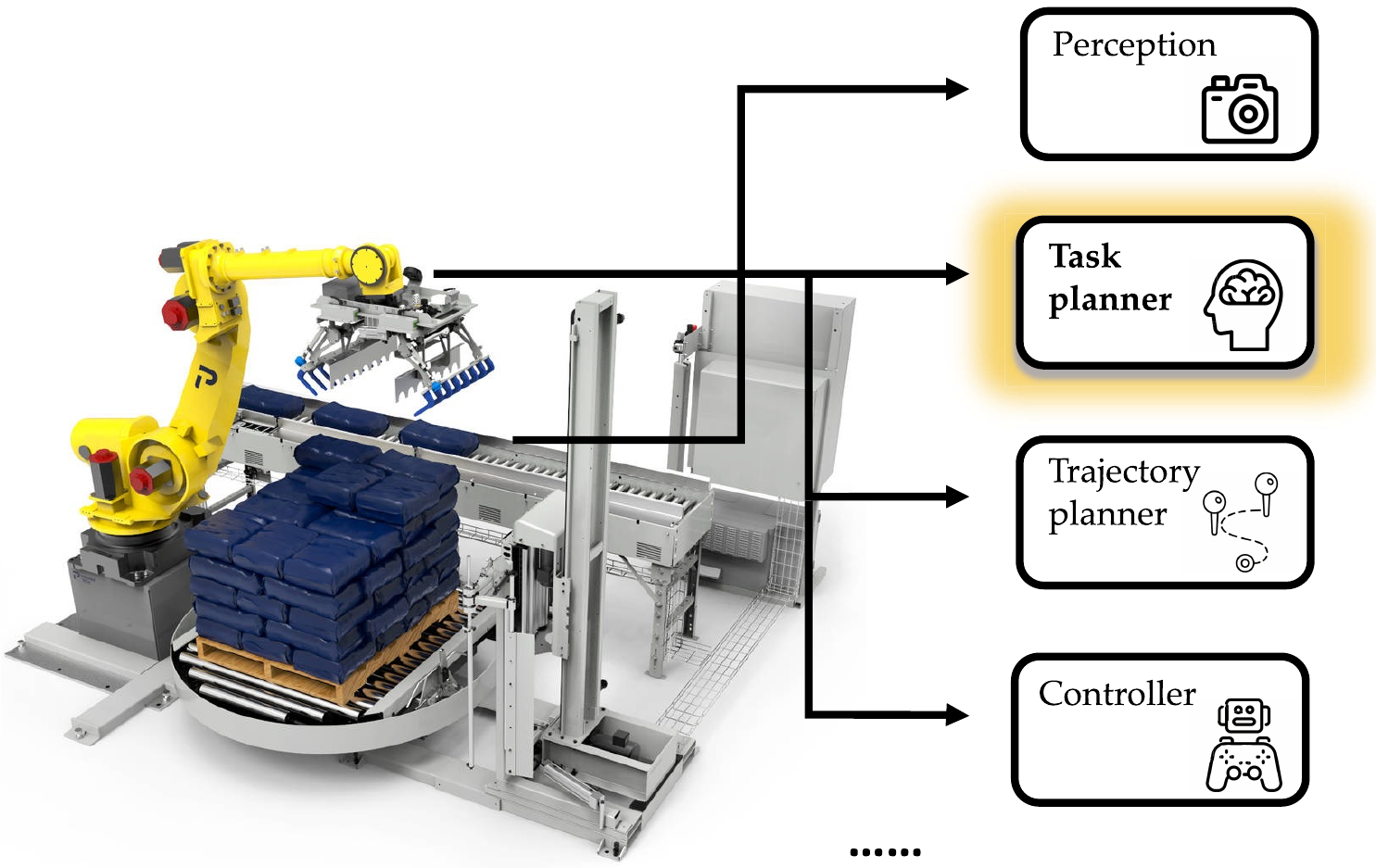}
	\caption{A typical robotic palletization system is composed of various modules, including perception, task planner, trajectory planner, controller, etc. Task planning is the main focus of our work.}
\label{fig:palletization_system}
\end{figure}

The demand for efficient palletization in warehouses and logistics centers has grown significantly in recent years. This increase is largely driven by the expansion of global trade and the rise of e-commerce, which require rapid and precise handling of goods. 
Traditional, manual methods of palletization are often unable to meet these demands due to their labor-intensive nature and potential for inconsistency.
Consequently, the development of robotic systems for palletization has become crucial. 
These robotic systems, as illustrated in Fig~\ref{fig:palletization_system},  are complex and composed of various integral components, including perception systems for item identification and localization, task planning modules for decision-making, trajectory planning and control modules for collision-free and precise execution, etc.
In this study, our focus is primarily on the task planning module of robotic palletization systems, which functions as determining the optimal picking order from a stream of boxes, deciding how each box should be rotated, and identifying the precise location for placing each box on the pallet, all in an online and dynamic fashion.

The task planning for robotic palletization in most industrial environments can be conceptually framed as a variant of the online 3D Bin Packing Problems (BPP), where the inventory of items is predetermined, yet their sequence of arrival remains unpredictable~\cite{wang2020robot}. 
Furthermore, in our work we address a more realistic and challenging problem setting that reflects common practice in robotic palletization systems by including a "buffer" area. This consideration allows the robot to utilize an auxiliary space for storing up to $N$ pending items, thereby expanding its operational capabilities beyond merely handling the immediate one.
While many research efforts have been made towards addressing online 3D BPP~\cite{karabulut2004hybrid, ha2017online, wang2019stable, wang2020robot}, most of the studies rely on hand-coded heuristic methods.
In contrast to these approaches, our work aims to statistically learn an optimal and physically feasible palletization task planner. 
While there are several existing works that use RL to solve online 3D BPP~\cite{zhao2021online, zhao2021learning, zhao2022learning}, these works are characterized by two primary limitations. 
Firstly, during policy training, these studies either neglect the aspect of packing stability~\cite{zhao2021online, zhao2021learning} or assess stability through heuristic-based analyses~\cite{zhao2022learning}, rendering the policies derived from such methods less viable for deployment in real-world systems where systematic uncertainties are present. 
Secondly, the problem settings in these works typically overlook the existence of the buffer area, thus potentially limiting their applicability and effectiveness in more complex scenarios.

In this work, we aim to provide an effective  solution to our specific online 3D BPP problem.
Due to the combinatorial nature of the action space of the problem, applying RL to solve online 3D BPP suffers from the problem of large action space. 
This challenge is further exacerbated with the introduction of the "buffer" area. 
One commonly used solution to the problem of RL with large action space is through "invalid action masking"~\cite{vinyals2017starcraft, berner2019dota, ye2020mastering}, which identifies and masks out the \textit{invalid} actions and directs the policy to exclusively sample valid actions during the learning phase. 
Our work adopts a similar approach, pinpointing valid actions as task plans that guarantee the stability of intermediate item stacks on the pallet under gravitational forces throughout the training process. 
Accurately estimating action masks (i.e., valid actions) for varying states during RL training is crucial, albeit challenging. 
To address this, we introduce a methodology that learns to estimate action masks through supervised learning (Section~\ref{sec:method:action_pruner}). 
Specifically, leveraging the image-like characteristics of observation and action mask data, we employ a semantic segmentation paradigm~\cite{chen2014semantic, long2015fully} to train the action masking model.  
The training dataset is compiled through an offline data collection phase, utilizing a physics engine to verify the stability of specific placements across various pallet configurations. 
Moreover, to ameliorate the potential issue of distribution shift inherent in our approach, we propose a DAgger-like framework~\cite{ross2011reduction} designed to iteratively refine the action masking model, thereby enhancing the RL policy's performance (Section~\ref{sec:method:iterative}).

To assess the efficacy of our proposed methodology, we conducted extensive experiments and compared our results with existing RL-based online 3D BPP solutions. 
Experimental results indicate that our proposed method outperforms other baselines and is more sampling-efficient. 
We also demonstrate the effectiveness of our proposed iterative framework through experimental evaluations. 
Lastly, our learned task planner is deployed in a real-world robotic palletizer to demonstrate the robustness and applicability of our approach in real-world settings.

%% file: related.tex
\section{RELATED WORK}
In this section, we provide a brief review of research on 3D Bin Packing Problems (BPP) under both the offline and online settings. 
\subsection{Offline 3D BPP}
The problem of 3D bin packing, a complex variant of the classical bin packing problem, involves efficiently packing objects of various sizes into a finite number of bins or containers with fixed dimensions in three dimensions. 
The offline version of the problem, where all information about the objects to be packed is known in advance, has been particularly challenging due to its NP-hard nature.
Early strategies were heavily reliant on heuristic methods, as initially demonstrated by~\cite{martello2000three} that extended 2D BPP algorithms to 3D context. 
This is followed by many research efforts that propose various heuristic and meta-heuristic algorithms to approximate the solutions~\cite{faroe2003guided, crainic2009ts2pack, kang2012hybrid}. 
More recently,~\cite{hu2020tap, zhang2021attend2pack} propose to use deep reinforcement learning (DRL) to learn the policy for offline 3D BPP. 
Our work differs from these works by focusing on online 3D BPP instead of offline version.

\subsection{Online 3D BPP}
In contrast to offline scenarios, online 3D bin packing problems involve making packing decisions without knowledge of future items, presenting unique challenges in achieving optimality and efficiency. 
Research in this domain has focused on developing adaptive heuristics and algorithms that can handle the sequential nature of item arrivals. 
Notably, Karabulut et al.~\cite{karabulut2004hybrid} proposes deep-bottom-left (DBL) heuristics and Ha et al.~\cite{ha2017online} extends ~\cite{karabulut2004hybrid} by employing a strategy that orders empty spaces using the DBL method and positions the current item into the earliest suitable space. 
~\cite{wang2019stable} introduces a Heightmap-Minimization technique aimed at reducing the volume expansion of items in a package when viewed from the loading perspective. 
Aside from heuristics-based methods, there are also research attempts utilizing DRL to solve the problem of online 3D BPP~\cite{zhao2021online, zhao2021learning}. 
DRL for online 3D BPP typically suffers from the problem of large action space, making the RL algorithms hard to optimize. 
~\cite{zhao2021online} address this challenge by implementing straightforward heuristics to reduce the action space, yet their demonstrations are confined to scenarios with limited discretization resolution. 
Conversely, another study by~\cite{zhao2021learning} seeks to alleviate this issue by devising a packing configuration tree that employs more intricate heuristics to identify a subset of feasible actions, marking a progression from their prior work. 
Our research is distinct from these approaches in two critical aspects. Firstly, the scenario we investigate is notably more complex and mirrors real-world conditions more closely. Specifically, we introduce a "buffer" area, enabling the agent to select any item from a queue of upcoming boxes and rotate it along any axis, thereby significantly expanding the action space. Furthermore, our model accounts for translational and rotational uncertainties that occur when a robot places a box onto the pallet—factors that are overlooked by~\cite{zhao2021learning, zhao2021online} but are prevalent in actual physical systems. 
Secondly, from a methodological perspective, our approach diverges from the reliance on varying heuristics to trim the action space. Instead, we introduce a supervised learning-based pipeline to develop an action masking mechanism.

%% file: framework.tex
\section{OUR PROPOSED APPROACH}
\label{sec:method}
In this section, we provide detailed description of our proposed method. Initially we specify the problem setting of the online 3D Bin Packing Problem (BPP) addressed in this study, as outlined in Section~\ref{sec:method:formulation}. Then  a comprehensive description of our action masking pipeline, along with its integration into reinforcement learning (RL), is presented in Section~\ref{sec:method:action_pruner}. Furthermore, we introduce the iterative adaptation of our proposed algorithm, presented in Section~\ref{sec:method:iterative}, illustrating the progressive refinement of our approach.

 \subsection{Problem Formulation}
 \label{sec:method:formulation}
 The online 3D BPP can be formulated as a Markov decision process (MDP), denoted by $\mathcal{M} = (\mathcal{S}, \mathcal{A}, \mathcal{P}, \mathcal{R})$, where $\mathcal{S}$ signifies the state space, $\mathcal{A}$ denotes the action space,  $\mathcal{P}$ represents the transition probability between states, and $\mathcal{R}$ encapsulates the real-valued rewards. 
The objective of RL is to find a policy, $\pi(a | s)$, that can maximize the accumulated discounted reward $J(\pi) = \mathbb{E} \left[ \sum_{t=0}^{\infty} \gamma^t R_t \right]$, where $\gamma$ represents the discounted factor. 
In the following, we details the components of the MDP in this study.

\paragraph{State} 
In the process of palletization, the policy $\pi$ evaluates two primary sources of information: the present configuration of the pallet and the dimensions of the forthcoming boxes.  
Consequently, we model the state at any given time $t$ as $s_t = \{ \mathcal{C}_t , \mathrm{d}_t \}$, where $\mathcal{C}_t$ denotes the current configuration of the pallet, and $\mathrm{d}_t$ encapsulates the dimensions of the forthcoming boxes. 
Following ~\cite{zhao2021online}, we employ a \textit{height map} to depict the pallet configuration. 
This involves discretizing the pallet into an $l_p \times w_p$ matrix, where each element signifies the discretized height at the corresponding location on the pallet. 
The variable $\mathrm{d}_t$ is conceptualized as a list of $N$ 3-dimensional tuples, with each tuple specifying the (\textit{length}, \textit{width}, \textit{height}) of a box. 
Here, $N$ represents the capacity of the "buffer" area, which is ascertained by the perception system and typically ranges between 3 and 5.

\paragraph{Action} 
At each timestep, the policy $\pi$ is tasked with executing three sequential decisions: selecting a box from the $N$ boxes in the upcoming queue (buffer area), rotating the box to the desired orientation, and positioning the box on the pallet. 
We restrict our consideration to axis-aligned orientations of the boxes, yielding $6$ possible orientations.
Furthermore, the front-left-bottom (FLB) corner of the box can be positioned in any grid cell among the $l_p \times w_p$ cells available on the pallet.  
Consequently, the action space is of a combinatorial nature, characterized by a dimensionality of $N \times 6 \times l_p \times w_p$, i.e., action $a \in \mathbb{R}^{N \times 6 \times l_p \times w_p}$. 
In a typical industrial context, with $N=5$ and $l_p = w_p = 25$ (at a 1-inch discretization resolution), this translates to an action space with $18,750$ dimensions. 
The extensive size of the action space introduces additional complexity to the RL problem, complicating the optimization process~\cite{dulac2015deep}.

\paragraph{Reward}
We conceptualize the accumulated reward as a measure of the \textit{space utilization} of the current pallet, contingent upon the \textit{stability} of the items positioned on it. 
The reward function $R$ is formally defined as follows:
\begin{equation}
    R(s_t) = \mathbb{1}(s_t) \cdot \frac{V_{total}}{V_{max}}
\end{equation}
Here, $\mathbb{1}(s_t)$ serves as an indicator function that validates the physical stability of the boxes on the pallet at state $s_t$. 
$V_{total}$ refers to the cumulative volume of the boxes currently situated on the pallet, while $V_{max}$ signifies the theoretical maximum volume the pallet can accommodate. 
Given considerations for safety during transport, it is assumed that the maximum height of the boxes on the pallet should not surpass $h_p = 25 $ inches, thus defining  $V_{max} = l_p \cdot w_p \cdot h_p = 25 ^3$.
Consequently, we can delineate a dense reward at each timestep as $r_d (s_t) = \mathbb{1}(s_t) \cdot \frac{V_{t}}{V_{max}}$, where $V_t$  denotes the volume of the most recently placed item.

\begin{figure*}[t]
\centering
   \includegraphics[width=\textwidth]{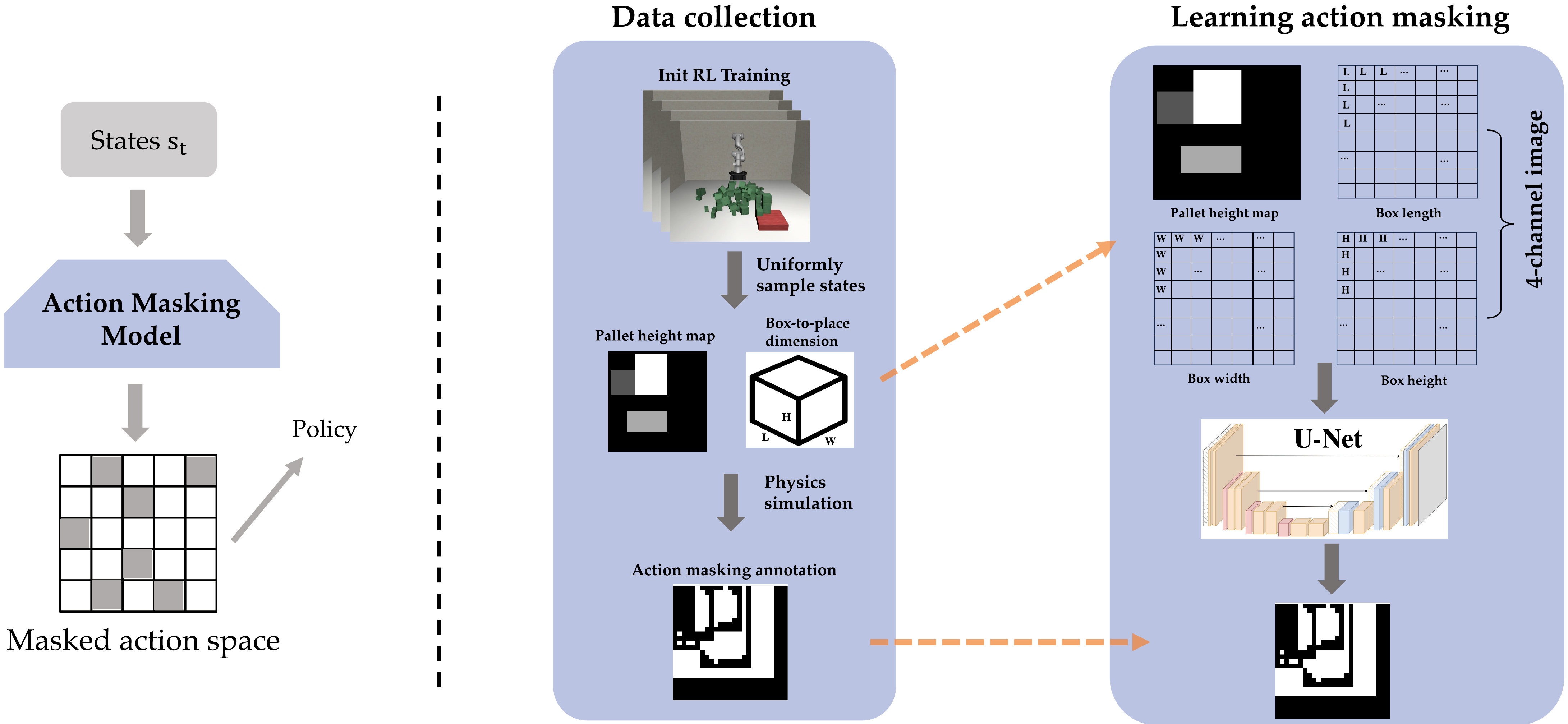}
  \caption{An overview of our action masking learning process. The methodology unfolds in three phases: data collection, for gathering relevant training data; learning the action masking model, where a U-net architecture learns to distinguish stable from unstable placements; and embedding the learned action masking model into RL training, integrating the model to dynamically reduce the action space and enhance RL optimization.}
  \label{fig:framework}
\end{figure*}

\subsection{Action Space Masking via Supervised Learning}
\label{sec:method:action_pruner}
The principal challenge in applying RL to address the online 3D BPP resides in the expansive dimensionality of the action space, which complicates the optimization of RL algorithms.  
To counteract this issue, we embrace the strategy of "invalid action masking" ~\cite{vinyals2017starcraft, berner2019dota, ye2020mastering}, which serves to alleviate the aforementioned challenge.
This section elucidates our methodology for developing the action masking model  through supervised learning. 
The fundamental premise behind action space masking is the exclusion of invalid placements that could culminate in unstable pallet configurations.
While prior studies have implemented similar strategies to exclude unstable placements and thereby simplify the learning process\cite{zhao2022learning, zhao2021online}, their approaches to action masking rely predominantly on heuristic algorithms. 
We posit that heuristic-based action masking exhibits several limitations: 
(i) The necessity for multiple hyperparameters in most action masking methods poses a challenge in tuning these parameters to accommodate diverse scenarios, such as variations in box sizes and weights. 
ii) Heuristic methods, particularly those involving the analysis of the center-of-mass (CoM), fail to account for uncertainties inherent in palletization execution, necessitating further elaboration. 
For instance, consider a heuristic that uses CoM analysis to prevent tipping by restricting box placements that significantly shift the CoM. 
However, this doesn't consider real-world uncertainties like slight placement inaccuracies due to robot sensor errors, leading to unexpected CoM shifts and potentially unstable configurations despite heuristic approval. 
This example underscores the limitations of CoM-based heuristics, highlighting their lack of accommodation for practical uncertainties in robot execution, which questions their reliability in real-world scenarios. 
In response, our work investigates the feasibility of adopting supervised learning to construct the action masking model. 
Figure~\ref{fig:framework} delineates the framework of our proposed methodology for learning action masking. 
Specifically, the learning process is segmented into the following phases:

\paragraph{Data Collection}
The action masking models eliminate unstable placements by learning to infer the stability of sampled box placements during RL training process. 
Specifically, we aim to identify a set of feasible placements, or action masks $g_t \in \mathbb{R}^{l_p \times w_p}$, based on the configurations of the pallet $\mathcal{C}_t$ and the (rotated) dimensions of the boxes $b_t$, formulated as $g_t = f_\theta(\mathcal{C}_t, b_t)$. 
The collection of data samples consisting of $(\mathcal{C}_t, b_t)$ pairs along with their corresponding $g_t$ is essential for the learning of the action masking model $f_\theta$. 
However, ensuring that the collected data accurately reflects the RL training distribution poses a significant challenge.
To address this, we initially employ a heuristic-based method to develop an initial RL policy $\pi_o$, from which we systematically sample and store $(\mathcal{C}_t, b_t)$ pairs for subsequent offline data collection.

In the data collection phase, our goal is to generate the corresponding annotations $g_t$ given the sampled RL states $(\mathcal{C}_t, b_t)$. 
Specifically, $g_t$ represents the placement stability on every discretized bin of the pallet for a given pallet state $\mathcal{C}_t$ and box dimension $b_t$. 
To achieve this, we leverage the physics simulation, i.e., MuJoCo~\cite{todorov2012mujoco}, to evaluate the stability of every potential placement on the pallet, thereby generating $g_t$. 
The dataset thus compiled, denoted as $\mathcal{D} = \{(\mathcal{C}_t, b_t, g_t)\}$, forms the basis for the subsequent learning stage of the action masking model.

\paragraph{Learning the Action Masking Model}
As illustrated in Figure~\ref{fig:framework}, $g_t$ functions as a binary image corresponding in size to $\mathcal{C}_t$, with pixels valued at $1$ indicating stable placements and those valued at $0$ signifying unstable ones. 
Leveraging the image-like properties of both the input and the output, we transform the learning challenge into a well-established computer vision task known as semantic segmentation~\cite{chen2014semantic, long2015fully, ronneberger2015u}. To this end, we generate three single-channel images mirroring the dimensions of $\mathcal{C}_t$, each channel encoding the length, width, and height of the box, respectively. These images are then combined with $\mathcal{C}_t$ to produce a 4-channel image, which serves as the input, with $g_t$ acting as the target output. This input-output pair is subsequently processed by a conventional U-net~\cite{ronneberger2015u} architecture for training purposes. The resulting neural network, regarded as the trained action masking model $f_\theta$, is subsequently integrated into the RL training regimen. 

\paragraph{Embedding the Learned Action Masking Model into RL Training}
We employ the developed action masking model to trim the action space during RL training, enhancing the efficiency of the learning process. 
At each timestep, with the pallet configuration $\mathcal{C}_t$ and the dimensions of the box to be placed $b_t$ given, the model $f_\theta$ identifies a viable subset $\hat{g}_t$ from the total available placements ($l_p \cdot w_p$ placements). 
The RL policy $\pi$ is then restricted to selecting placements solely from this feasible subset $\hat{g}_t$. 
This strategy effectively narrows the action space, significantly facilitating the RL training. 
We demonstrate the effectiveness of the action masking model for RL training in Section~\ref{sec:exp}.

\subsection{Iterative Action Masking for RL Training}
\label{sec:method:iterative}
While being effective, the learned action masking model can still be inaccurate in certain states, particularly for the out-of-distribution (OOD) states encountered during the RL training process. 
Such discrepancies arise because during RL training the agent might explore states that diverge significantly from those within the training dataset $\mathcal{D} = {(\mathcal{C}_t, b_t, g_t)}$, leading to a misalignment between the distributions of the training data and the actual scenarios the model faces. 

Inspired by DAgger~\cite{ross2011reduction}, we introduce an iterative enhancement pipeline for refining the learning of the action masking model. 
This process begins with the application of a heuristic-based method for initial action masking during RL training, yielding the initial policy $\pi_0$. 
Concurrently, we systematically collect state samples during $\pi_0$ training, assembling dataset $\mathcal{D}_0$ via offline collection as detailed in Section~\ref{sec:method:action_pruner}. 
Utilizing $\mathcal{D}_0$, we then train the neural network-based action masking model $f_\theta^0$, as depicted in Section~\ref{sec:method:action_pruner}. 
This model, $f_\theta^0$, is subsequently integrated into a new RL training cycle to derive an enhanced policy $\pi_1$. 
This cycle of training, data collection to form $\mathcal{D}_i$, and model updating to $f_\theta^i$ is repeated, each iteration aiming to embed the updated action masking model in the subsequent RL training phase. 
This iterative process continues until the RL policies cease to show significant improvements, thereby optimally aligning the action masking model with the encountered state distributions. 
The pseudo code of the proposed method is presented in Algorithm~\ref{algo:iterative}.
\begin{algorithm}[t]
    \caption{Iterative action masking model learning for RL training.} 
    \begin{algorithmic}
        \State RL training with heuristics-based action masking to obtain initial policy $\pi_0$.
        \State Collect initial dataset $\mathcal{D}_0$ by uniformly sampling from $\pi_0$ training.
        \State Initialize $\mathcal{D} \leftarrow \mathcal{D}_0$.
        \State Train action masking model $f_\theta ^ 0$ with $\mathcal{D}$.
        \For {i=1,...,N}
            \State RL training with $f_\theta ^ i$ to obtain policy $\pi_i$.
            \State Collect dataset $\mathcal{D}_i$ by uniformly sampling from $\pi_i$ training.
            \State Aggregate datasets: $\mathcal{D} \leftarrow \mathcal{D} \cup \mathcal{D}_i$
            \State Train $f_\theta ^ i$ with $\mathcal{D}$.
        \EndFor
        \State \Return best $\pi_i$ on testing.
    \end{algorithmic}
    \label{algo:iterative}   
\end{algorithm}

%% file: experiment.tex
\section{EXPERIMENTAL VALIDATIONS}
\label{sec:exp}
In this section, we present a thorough quantitative analysis of our proposed approach. Our evaluation is structured to address the following key questions: 1) Does the method outlined in Section~\ref{sec:method:action_pruner} successfully learn to identify the feasible subset of placements? 
2) Does the integration of the learned action masking model enhance the RL training process, culminating in a superior policy? 
3) Does the iterative action masking learning strategy described in Section~\ref{sec:method:iterative} contribute to further enhancements in the RL agent's ultimate performance?

\begin{figure}[h]
	\centering
	\includegraphics[width=0.8\linewidth]{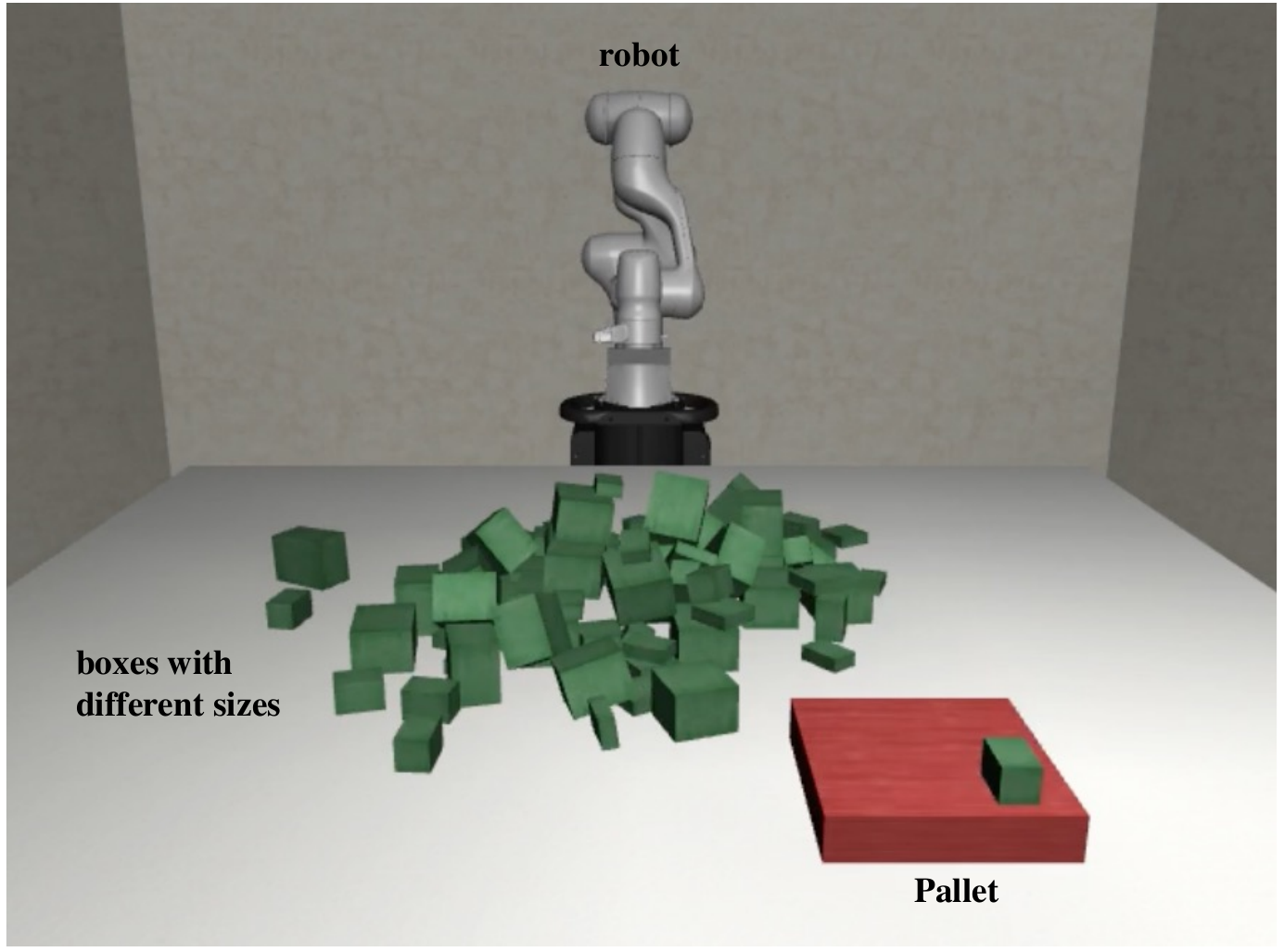}
	\caption{Visualization of our simulated palletization environment in MuJoCo~\cite{todorov2012mujoco}. Although 80 boxes are displayed for illustrative purposes, the robot is programmed to perceive and interact with only $N$ boxes within the buffer area. The arrangement of the boxes is randomized and unknown, shuffled anew for each RL episode.}
\label{fig:exp_setup}
\end{figure}

\subsection{Experimental Setup}
\label{sec:exp:setup}
To explore the task planning challenges associated with the palletization problem, we developed a simulated palletization task in MuJoCo~\cite{todorov2012mujoco}, reflecting a realistic logistic scenario, as depicted in Figure~\ref{fig:exp_setup}.  
This simulation environment features 80 boxes of 5 distinct sizes, with dimensions (in inches) of $10 \times 8 \times 6$, $9 \times 6 \times 4 $, $6 \times 6 \times 6 $, $6 \times 4 \times 4 $, $4 \times 4 \times 4 $.
The pallet is specified to have dimensions of $25 \times 25$ inches, with a stipulation that the maximum height of stacked boxes cannot exceed 25 inches.  
For our experiments, we applied a discretization resolution of 1 inch to both the boxes and the pallet. 

At each planning step, the task planner receives the current pallet configuration, represented as a height map, and the dimensions of $N$ forthcoming boxes in the buffer, then generates a task plan.  
This plan involves selecting one of the $N$ boxes, orienting it as desired, and placing it on the pallet, in accordance with the methodology described in Section~\ref{sec:method:formulation}.  
To expedite the learning process, the actual 'pick-rotate-place' actions by the robot are bypassed; instead, the chosen box is immediately positioned at the planner's determined goal pose.  
To account for uncertainties inherent in physical execution, each box placement is subjected to random translational noise on the xy plane $\delta_t \sim \mathcal{N}(\textbf{0}, \textbf{0.05})$ and rotational noise around the z-axis $\delta_r \sim \mathcal{N}(0, 5)$, measured in inches and degrees, respectively. 
The sequence in which the 80 boxes are presented to the policy is randomized and unknown at the start of each RL training episode, and the weights of the boxes are also randomized to further enhance the simulation's fidelity to real-world variability. 

\begin{figure}[h]
	\centering
	\includegraphics[width=\linewidth]{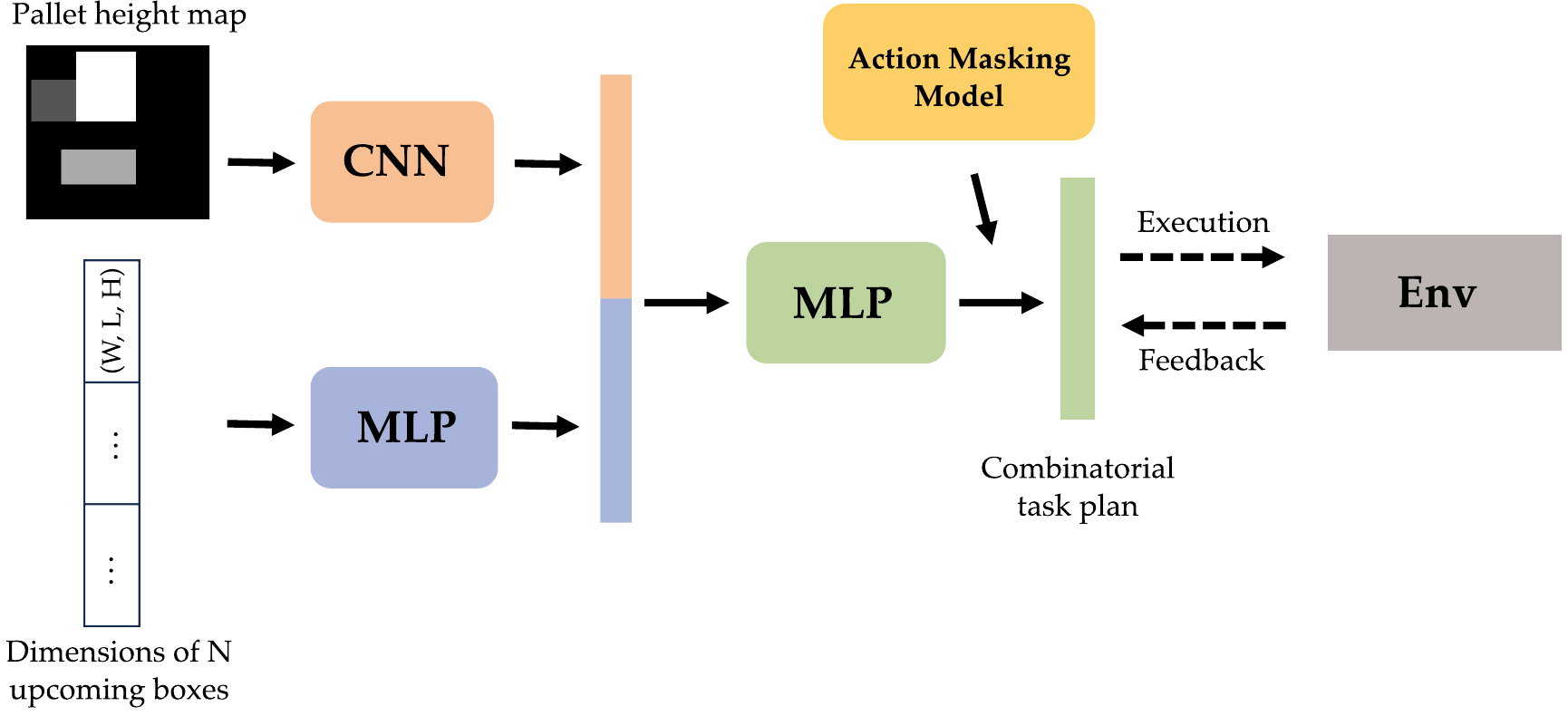}
	\caption{The policy network architecture adopted in our study. We use a CNN to encode the height map and a MLP to encode the dimensions of forthcoming boxes. The resulting embeddings are concatenated and serve as the input to the policy network. The  action masking model, if it exists, helps the policy network ignore infeasible actions during learning.}
\label{fig:policy_arch}
\end{figure}

\subsection{Implementation Details and Baselines}
\label{sec:exp:baselines}
Our objective is to validate the efficacy of the learned action masking model within the RL training framework. To this end, we implemented two baseline methodologies, \texttt{NoMask} and \texttt{HeuristicsMask}, for comparison against our proposed \texttt{LearnedMask} method, detailed in Section~\ref{sec:method:action_pruner}. The implementation specifics are outlined as follows: 
\paragraph{\texttt{NoMask}}
In the \texttt{NoMask} approach, action masking is not applied during the RL training process. Although this method can be utilized with various RL algorithms, we have chosen to implement PPO~\cite{schulman2017proximal} for all experiments conducted in this study. During the training, the policy receives as input the discretized pallet height map alongside the dimensions of $N$ forthcoming boxes, subsequently generating a combinatorial task plan. To process this information, a Convolutional Neural Network (CNN) is used to encode the 2D height map into a vector, while a Multi-Layer Perceptron (MLP) encodes the dimensions of the incoming boxes. These vectors are then concatenated and supplied to the policy as input. The architecture of the policy is depicted in Figure~\ref{fig:policy_arch} for further clarification.

\paragraph{\texttt{HeuristicsMask}}
The \texttt{HeuristicsMask} configuration is similar to \texttt{NoMask}, with the distinction of incorporating a heuristics-based action masking strategy during RL training. Specifically, we employ the heuristic criteria outlined in Zhao et al.~\cite{zhao2021online} to determine the feasible subset $\hat{g}_t$ at each timestep. According to these criteria, a placement location is deemed feasible if it satisfies any of the following conditions: 1) more than $60\%$ of the base area of the box to be placed, along with all four of its bottom corners, receive support from the boxes already on the pallet; 2) more than $80\%$ of the base area and at least three of the four bottom corners are supported; or 3) more than $95\%$ of the base area is supported.

\paragraph{\texttt{LearnedMask}}
\texttt{LearnedMask} (Ours) also shares the same settings as \texttt{NoMask}, except that our proposed learning-based action masking is performed during RL training, as described in Section~\ref{sec:method:action_pruner}.

\subsection{Experimental Results}
\label{sec:exp:results}
In this section, we provide evaluation results under different experimental settings, aiming to address the three questions posed at the outset of Section~\ref{sec:exp}. 

\subsubsection{Learned Action Masking Model as Accurate Stability Predictor}
A fundamental approach to assessing the effectiveness of the learned action masking model involves evaluating the accuracy of its predictions. 
As outlined in Section~\ref{sec:method:action_pruner}, the action masking model  functions as an estimator, predicting the stability of specific placements and essentially operates as a semantic segmentation model. 
Therefore, we employ the standard Intersection-over-Union (IoU) metric to gauge the action masking model's performance in stability prediction. 
The performance metrics for both our learned action masking model and the heuristic-based action masking, as introduced in Section~\ref{sec:exp:baselines}, are reported for comparison.
The heuristic-based action masking achieves an IoU of $76.6\%$, whereas our learned action masking attains an IoU of $89.2\%$, markedly surpassing the heuristic approach. This result demonstrates that the learned action masking model can provide a more accurate action space mask for policy sampling than the heuristic-based method, potentially enhancing RL training outcomes.

\subsubsection{RL Policy Performance Improvement with Learned Action Masking}
\begin{figure}[h]
	\centering
	\includegraphics[width=0.95\linewidth]{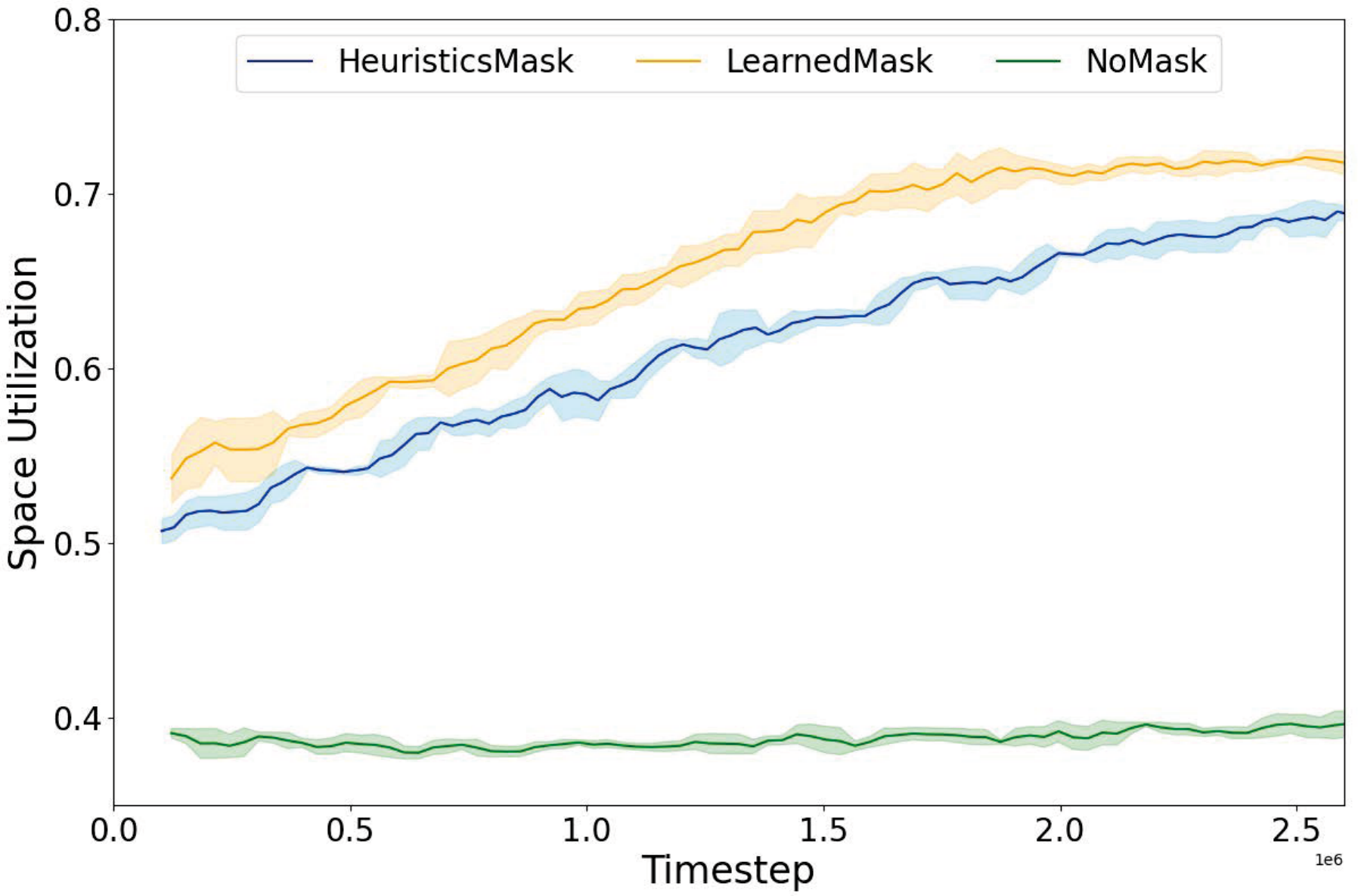}
	\caption{Learning curve of the three methods when buffer size $N=1$. Results are averaged over 5 random seeds. Our method (\texttt{LearnedMask}) converges faster and achieves better space utilization compared to the baseline methods.}
\label{fig:learning_curve}
\end{figure}

Intuitively, a precisely calibrated action masking model can significantly reduce the exploration space inherent in the original RL problem, thereby simplifying the optimization process.  
To empirically verify this, we conducted experiments by comparing the policy efficacy of our proposed \texttt{LearnedMask} model against two baselines: \texttt{NoMask} and \texttt{HeuristicsMask}, as elaborated in Section~\ref{sec:exp:baselines}. 
We adopted the \textit{space utilization} as defined in Section~\ref{sec:method:formulation} as the reward. 
Figure~\ref{fig:learning_curve} shows the performance of the three methods under the setting when the buffer size $N = 1$ as training proceeds, across $5$ random seeds. 
Observations reveal that both \texttt{LearnedMask} and \texttt{HeuristicsMask} significantly surpass \texttt{NoMask} in performance, thereby affirming the hypothesis that optimization within the original action space of RL is inherently challenging. Action masking thus serves to effectively narrow the problem domain, substantially enhancing policy performance. 
Moreover, \texttt{LearnedMask} not only converges more rapidly but also achieves superior performance relative to \texttt{HeuristicsMask}, indicative of the premise that a more precise action masking model correlates with improved policy learning.  

In a bid to comprehensively assess policy performance across variable configurations, we tabulated the average space utilization metrics for all three methods under varying buffer sizes, specifically, $N = 1, 3, 5$. These results were derived from executing the corresponding policy across 20 iterations and computing the mean reward from these episodes. Notably, during these test episodes, the arriving order of the boxes is random and unknown to the policy, identical to the training phase. The outcomes of this analysis are encapsulated in Table~\ref{tab:performance}. As depicted, \texttt{LearnedPrune} consistently outperforms \texttt{HeuristicsPrune} across all experimental conditions, thereby evidencing the  robustness of our algorithm. Additionally, an increase in $N$ correlates with an enhancement in policy performance, a logical deduction considering the expanded decision-making vista afforded by a larger buffer. Nonetheless, in practical applications, the feasible value of $N$ is constrained by various factors, including the limitations of the perception system and physical spatial constraints. Within the context of most industrial palletization scenarios, a buffer size of $N = 5$ represents a pragmatic upper limit.

\begin{table}[t]
\centering
\caption{Average space utilization of the obtained policies under different buffer size $N$.}
\begin{tabular}{@{}lccc@{}}
\toprule
Buffer size & $N =1$ & $N =3$ & $N =5$ \\ \midrule
\texttt{NoMask} & 38.8\% & 37.6\% & 38.1\% \\
\texttt{HeuristicsMask} & 69.4\% & 71.9\% & 72.7\% \\
\texttt{LearnedMask}   & \textbf{72.1\%} & \textbf{75.1\%} & \textbf{76.2\%} \\  \bottomrule
\end{tabular}
\label{tab:performance}
\end{table}

\subsubsection{Iterative Policy Improvements with Iterative Action Masking Learning}
\begin{figure}[h]
	\centering
	\includegraphics[width=0.8\linewidth]{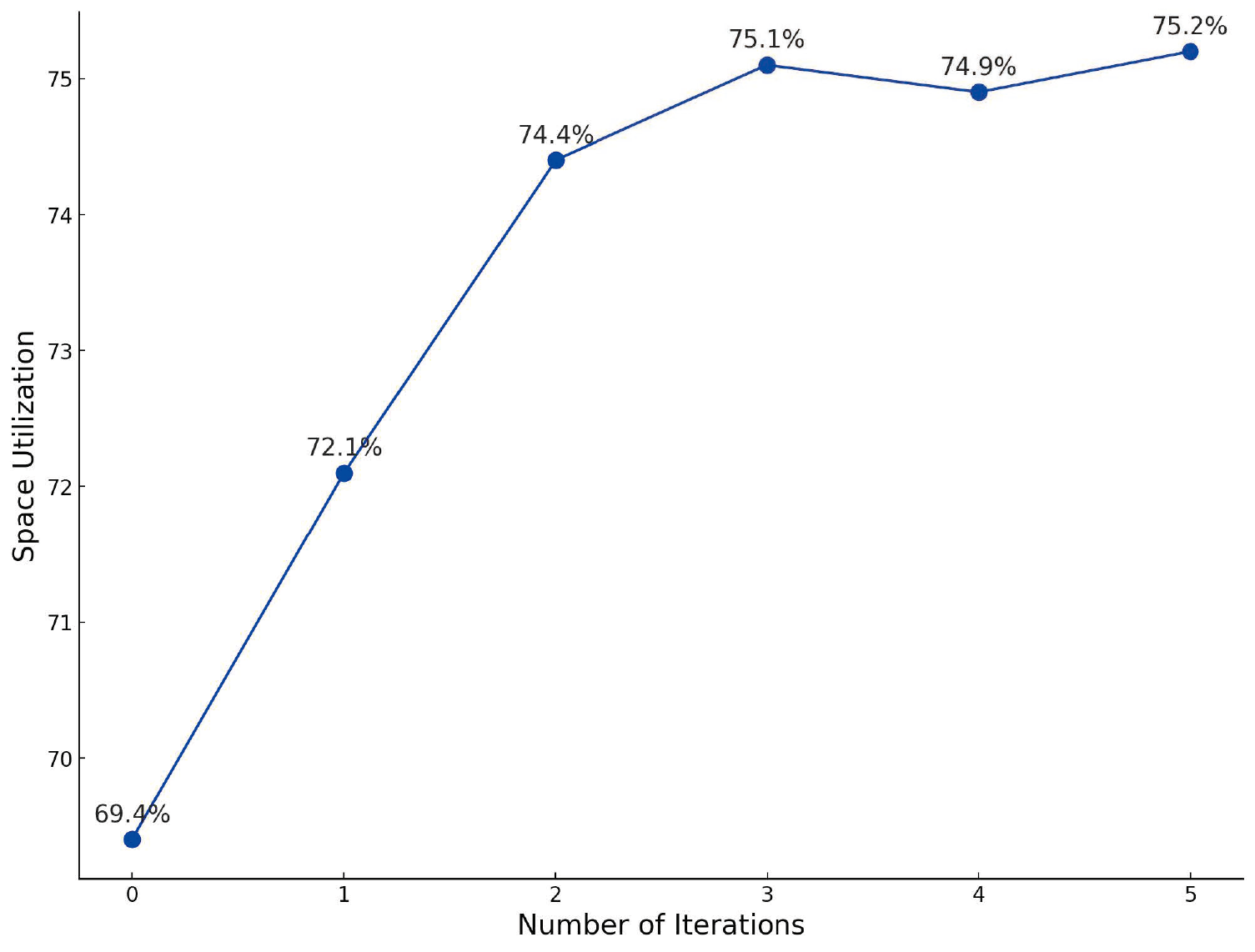}
	\caption{The policy performance along with the number of iterations $T$. The first two iterations ($T = 2, 3$) significantly improves the performance, while the incremental improvement in performance begins to diminish in later iterations ($T = 4, 5$).}
\label{fig:iterative_performance}
\end{figure}

Here we aim to investigate if the iterative action masking learning algorithm in Section~\ref{sec:method:iterative} can further improve the performance of RL policy over the iterations. 
Figure~\ref{fig:iterative_performance} delineates the relationship between policy performance and the number of iterations, denoted by $T$. 
Specifically, $T = 0$ denotes the baseline scenario of RL training underpinned by heuristic action masking, represented as \texttt{HeuristicsMask}, while $T = 1$ signifies the employment of \texttt{LearnedMask}. 
Through empirical investigation, we observed a notable enhancement in performance during the initial iterations ($T = 2, 3$), with improvements of $1.0\%$ and $0.3\%$ over their immediate predecessors, respectively.  
However, the policy performance "saturates" in subsequent iterations ($T = 4, 5$). 
We hypothesize that this phenomenon is attributed to the substantial mitigation of the initial distribution \textit{misalignment} issue within the first two iterations ($T = 2, 3$), which directly contributes to the incremental improvements of the derived policies. 
Consequently, beyond this point, the performance is no longer impeded by the \textit{misalignment} challenge, rendering additional iterations ineffectual in further elevating the policy's performance. 

\subsection{Real-World Deployment}

\begin{figure}[h]
	\centering
	\includegraphics[width=\linewidth]{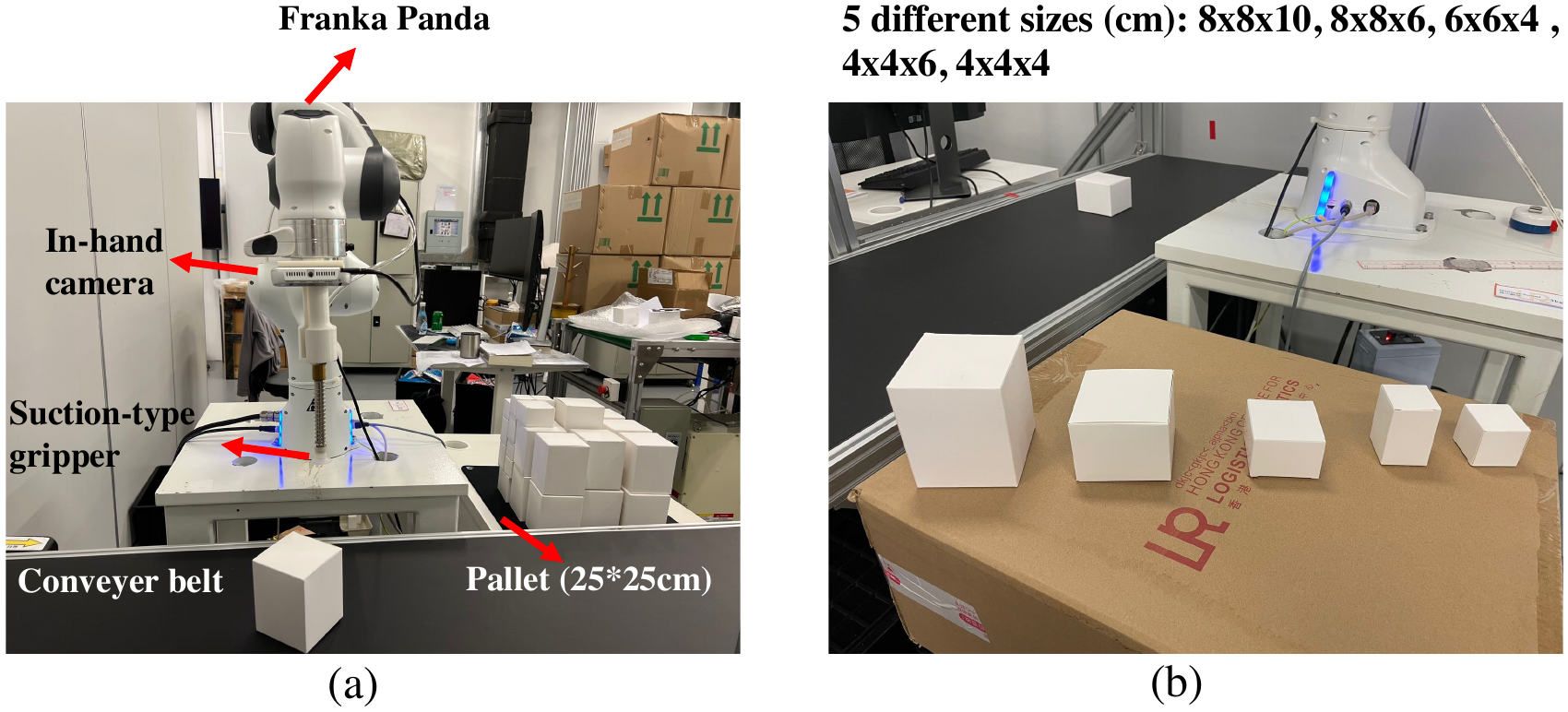}
	\caption{(a) Real-world experiment setup. (b) The boxes (of 5 different sizes) used in the real-world experiment.}
\label{fig:real_system}
\end{figure}

We further tested our RL-based palletization task planner in a real-world physical system. Due to limitations in space and hardware, it was not feasible to implement our algorithm on an industrial-scale robot handling large boxes. Therefore, we utilized a Franka Panda robotic arm as a prototype for our physical system setup, depicted in Figure~\ref{fig:real_system}(a). The robot is equipped with a suction-type gripper for box manipulation and an Intel Realsense RGB-D camera mounted in-hand for box detection and localization on the conveyor. Space constraints restrict the camera's perception to a single box at a time, resulting in an upcoming box count $N=1$. The boxes, showcased in Figure~\ref{fig:real_system}(b), come in 5 different sizes and contain various items to simulate the diversity in weights found in real palletization tasks.

\begin{figure}[h]
	\centering
	\includegraphics[width=0.9\linewidth]{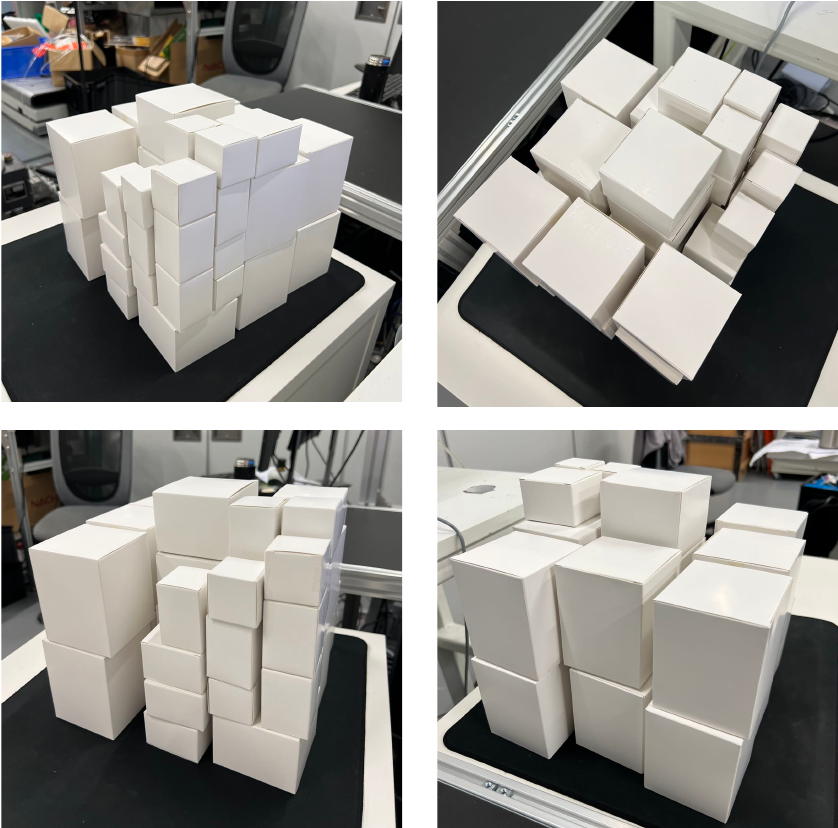}
	\caption{Final pallet configuration from different viewpoints. The resulting pallet is compact and stable, demonstrating the effectiveness and robustness of the learned task planner.}
\label{fig:final_pallet}
\end{figure}

Our methodology involved training the task planner in a simulated environment tailored to the physical system's specifications regarding pallet and box dimensions. This trained planner was then directly applied to the real-world setup. The robotic system demonstrated proficiency by successfully stacking 35 boxes on the pallet, achieving a space utilization rate of $72.0\%$ in a single episode. The final pallet configuration, viewed from different angles, is presented in Figure~\ref{fig:final_pallet}. The ability to stack 35 boxes while maintaining compactness and stability showcases the effectiveness and robustness of our learned palletization planner policy. A demo video of the palletization process can be accessed at \href{https://drive.google.com/file/d/1z-Q59Nk5q8In_WGsZVix7qozkqqEymDb/view?usp=sharing}{here}.


%% file: conclusion.tex
\section{CONCLUSION AND FUTURE WORK}

This study introduced a novel reinforcement learning-based approach for robotic palletization task planning, emphasizing the significance of iterative action masking learning to manage and prune the action space effectively. Our methodology combines the precision of supervised learning with the adaptive capabilities of reinforcement learning, showcasing substantial improvements in both the efficiency and reliability of task planning for robotic palletization. The experimental validations, both in simulated environments and real-world deployments, have demonstrated the enhanced learning efficiency and operational performance of our proposed method.

A major limitation of our current study is the lack of integration between the task planning phase and the generation of collision-free trajectories for robotic execution. Specifically, our methodology does not account for the ease with which robots can execute the generated plans without encountering physical obstacles. Merging the task planning process with trajectory planning to produce plans that are not only efficient but also physically feasible for robots to execute without collision presents a promising avenue for future research.